%% file: main.tex
\documentclass[10pt,twocolumn,letterpaper]{article}

\usepackage[pagenumbers]{iccv} 

\input{preamble}

\definecolor{iccvblue}{rgb}{0.21,0.49,0.74}
\usepackage[pagebackref,breaklinks,colorlinks,allcolors=iccvblue]{hyperref}

\def\paperID{12} 
\def\confName{ICCV}
\def\confYear{2025}

\title{
cubic: CUDA-accelerated 3D Bioimage Computing
}

\author{Alexandr A. Kalinin \qquad Anne E. Carpenter \qquad Shantanu Singh\\
Broad Institute of MIT and Harvard\\
Cambridge, MA 02142, USA\\
{\tt\small akalinin@broadinstitute.org}
\and
Matthew J. O’Meara\\
University of Michigan Medical School\\
Ann Arbor, MI 48109, USA\\
{\tt\small maom@umich.edu}
}

\begin{document}
\maketitle
\input{sec/0_abstract}

\input{sec/1_intro}
\input{sec/2_methods}
\input{sec/3_results}
{
    \small
    \bibliographystyle{ieeenat_fullname}
    \bibliography{main}
}

\end{document}

%% file: preamble.tex
\usepackage{listings}          
\usepackage{xcolor}            

\lstdefinestyle{custom}{
  language        = Python,
  basicstyle      = \footnotesize\ttfamily,
  keywordstyle    = \color{blue},
  commentstyle    = \color{gray},
  stringstyle     = \color{teal},
  showstringspaces= false,
  tabsize         = 4,
  columns         = fullflexible,
  keepspaces      = true,
  frame           = single,
  xleftmargin     = 2em,
  framexleftmargin= 1.6em
}

%% file: sec/0_abstract.tex
\begin{abstract}

Quantitative analysis of multidimensional biological images is useful for understanding 
complex cellular phenotypes and accelerating advances in biomedical research. As modern microscopy generates ever-larger 2D and 3D datasets, existing computational approaches are increasingly limited by their scalability, efficiency, and integration with modern scientific computing workflows.
Existing bioimage analysis tools often lack application programmable interfaces (APIs), do not support graphics processing unit (GPU) acceleration, lack broad 3D image processing capabilities, and/or have poor interoperability for compute-heavy workflows.
Here, we introduce \emph{cubic}, an open-source Python library that addresses these challenges by augmenting widely-used SciPy and scikit-image APIs with GPU-accelerated alternatives from CuPy and RAPIDS cuCIM.
\emph{cubic}'s API is device-agnostic and dispatches operations to GPU when data reside on the device and otherwise executes on CPU—seamlessly accelerating a broad range of image processing routines.
This approach enables GPU acceleration of existing bioimage analysis workflows, from preprocessing to segmentation and feature extraction for 2D and 3D data.
We evaluate \emph{cubic} both by benchmarking individual operations and by reproducing existing deconvolution and segmentation pipelines, achieving 
substantial speedups while maintaining algorithmic fidelity.
These advances establish a  robust foundation for scalable, reproducible bioimage analysis that integrates with the broader Python scientific computing ecosystem including other GPU-accelerated methods, enabling both interactive exploration and automated high-throughput analysis workflows.
\emph{cubic} is openly available at 
\url{https://github.com/alxndrkalinin/cubic}.
\end{abstract}

%% file: sec/1_intro.tex
\section{Introduction}
\label{sec:intro}

The growing scale and complexity of biological experiments—particularly in 
drug discovery, functional genomics, and systems biology—have elevated the importance of 
quantitative bioimage analysis~\cite{chandrasekaran2021image,bagheri2022new,ia2023s}.
Advances in microscopy now enable high-throughput imaging 
of single cells, organoids, and tissues in various imaging modalities and conditions, generating 
terabytes of multidimensional data~\cite{lukonin2021organoids,bagheri2022new,balasubramanian2023imagining}.
Extracting information from these rich  datasets is essential for identifying phenotypic signatures,
quantifying cellular states, and linking molecular perturbations to observable
outcomes~\cite{caicedo2016applications,way2022morphology}.
As a result, computational workflows for image-based profiling are becoming increasingly essential
to modern biological discovery and translational research.

Despite this central role, existing bioimage analysis pipelines face significant bottlenecks. 
While there are robust and open-source image analysis platforms such as ImageJ/Fiji~\cite{schneider2012nih,schindelin2012fiji} and
CellProfiler~\cite{stirling2021cellprofiler4} that have made
image processing and morphological profiling broadly accessible to the community,
these frameworks are challenging to scale to modern workloads. First, they tightly couple their graphical interfaces and APIs,
making it challenging to integrate them into automated high-throughput workflows that employ
modern scientific computing and machine learning technology stack, and second, their architecture makes it difficult to leverage GPU acceleration.

By contrast, general purpose image processing libraries like OpenCV~\cite{opencv_library} and
scikit-image~\cite{vanderwalt2014scikit} offer programmatic flexibility via Python interfaces. Still, 
these have limited ability to handle the large-scale, high-dimensional datasets
generated by modern imaging techniques.
The shift toward 3D imaging, time-lapse data, and high-content screens exacerbates limitations
in computational efficiency and scalability, making processing pipelines slow, fragmented,
or difficult to integrate with machine learning and data science workflows.

Over the last decade, the increasing availability of graphics processing units (GPUs) has enabled end-to-end training of complex neural network architectures on terabyte-scale
image datasets~\cite{abadi2016tensorflow,paszke2019pytorch}.
However, it has not translated into similar gains for classical image processing in bioimage analysis.
While deep learning frameworks use GPU acceleration to speed up model training,
they typically focus on implementing deep learning-specific differentiable modules and operations.

This leaves many commonly used traditional bioimage analysis routines locked to CPU workflows,
with limited device flexibility and cumbersome APIs for large-scale data movement and hybrid pipelines.
GPU-accelerated solutions are available for specific tasks or as separate plugins, but they often rely on different back-ends, involve complex
installation procedures, and expose narrow, custom interfaces that do not always offer
interoperability with the wider modern scientific computing and data analysis stack.
Thus, researchers navigate a fragmented software landscape, assembling ad-hoc scripts and file conversions instead of concentrating on the biological questions at hand.
While libraries mirroring scientific-Python APIs now provide GPU backends~\cite{cupy,cucim}, integrating them typically demands explicit device management—tracking data placement and routing each call to the matching CPU or GPU implementation.

To address these challenges, we introduce \emph{cubic}, an open-source Python library for morphological 
analysis of multidimensional bioimages that provides a minimal-code-change interface for
the device-agnostic execution with optional GPU acceleration.
Specifically, \emph{cubic} uses the CUDA-accelerated libraries CuPy and cuCIM to 
provide fast, efficient implementations of common image processing operations,
including deconvolution, segmentation, and feature extraction from large 3D microscopy datasets.
\emph{cubic} is designed to be fully compatible with the Python scientific computing ecosystem, as it 
mirrors SciPy and scikit-image APIs, allowing users to leverage familiar functions and
reducing the changes required to adapt existing codebases.
It also supports zero-copy data exchange with PyTorch~\cite{paszke2019pytorch}, enabling GPU-resident image processing operations to feed directly into deep learning models without extra
memory copies or performance overhead.
By transparently leveraging GPU acceleration when available, while remaining fully functional on CPU-only 
systems, \emph{cubic} bridges the gap between ease of use, scalability, and computational 
performance, enabling robust and reproducible morphometric workflows for the next 
generation of bioimage analysis applications.
\emph{cubic} is open-source and is available at 
\url{https://github.com/alxndrkalinin/cubic}.

\section{Related work}
\label{sec:related_work}

\subsection{Traditional bioimage analysis tools}

The bioimage analysis ecosystem is rich with widely used, mature platforms that have 
shaped the field~\cite{haase2022hitchhiker}.
ImageJ/Fiji~\cite{schneider2012nih,schindelin2012fiji} offer a vast plugin
ecosystem and user-friendly graphical user interface (GUI) for both 2D and some 3D processing tasks.
While widely adopted, it requires scripting macros for large-scale processing, making interoperability 
with the Python scientific computing stack challenging.
CellProfiler~\cite{stirling2021cellprofiler4} is implemented in Python and enables repeatable, modular
pipelines for segmentation and feature extraction, and supports headless batch operation.
However, it remains entirely CPU-bound and can be prohibitively slow for large volumetric images. 
Tools like ilastik and QuPath cater to interactive segmentation and classification tasks.
Ilastik~\cite{kreshuk2019ilastik} provides user-friendly pixel/object classification using machine learning,
with optional GPU acceleration only within deep learning modules.
QuPath~\cite{bankhead2017qupath}, designed for whole-slide pathology, now includes GPU support via
PyTorch~\cite{paszke2019pytorch}, but remains limited in core morphometric pipelines.
While these packages excel in usability and community support, they usually lack large 3D
image analysis capabilities and GPU acceleration.
Moreover, implementation of image analysis operations in interactivity-first tools is usually tightly
coupled to the GUI, making it difficult to assemble robust pipelines that involve other tools
from the modern scientific Python stack~\cite{munoz2025cp_measure}.

More recently, napari~\cite{napari2019,chiu2022napari} has emerged as a Python-native, multi-dimensional image viewer that also provides a headless API and a rich plugin ecosystem exposing SciPy and scikit-image~\cite{witz2024napari} APIs. While napari’s programmatic interface enables both interactive and scriptable workflows within a single framework, it still relies on plugin registration and viewer-centric constructs even when run headlessly and does not aim to natively provide GPU support for available image analysis operations.

OpenCV~\cite{opencv_library} and scikit-image~\cite{vanderwalt2014scikit} are popular libraries for image processing
and computer vision that offer programmatic flexibility and implement a wide range of algorithms.
OpenCV, while providing a Python interface, primarily focuses on 2D image analysis.
scikit-image is built on NumPy~\cite{harris2020array} and SciPy~\cite{virtanen2020scipy},
making it easy to integrate into the scientific Python stack.
For example, CellProfiler v4~\cite{stirling2021cellprofiler4} itself packaged most of its core image-processing routines into scikit-image.
Similarly, many end‐to‐end bioimage analysis pipeline frameworks~\cite{rasse2020opsef,palla2022squidpy,comolet2024scalefex,atgu2024microscopytools,olafsson2025spacr} leverage scikit-image routines under the hood to perform image preprocessing, segmentation, feature extraction, and analysis.
However, scikit-image itself is CPU-bound and does not natively support GPU acceleration.

\subsection{GPU-accelerated frameworks and tools}

Specialized GPU-accelerated plugins for Imagej/Fiji and CellProfiler exist—such as 
AutoDeconJ~\cite{autodeconj} for 3D light‑field deconvolution in ImageJ—but these remain
task-specific and not integrated into the broader scientific computing environment.
OpenCL-based CLIJ/CLIJ2~\cite{haase2020_clij,vorkel2020_gpu_macro} brings hundreds of classical 2D and 3D GPU-accelerated
image operations into ImageJ/Fiji.
However, CLIJ is also not natively Python-based and primarily operates within
the ImageJ/Fiji ecosystem.
clEsperanto and its Python binding pyclesperanto~\cite{pyclesperanto2023} address this issue
by exposing CLIJ2 operations to Python users, enabling GPU-accelerated image processing
in a device-agnostic manner.
However, pyclesperanto implements a limited range of bioimage analysis operations with 
an interface specific to CLIJ2.

Deep learning frameworks such as PyTorch~\cite{paszke2019pytorch} and
TensorFlow~\cite{abadi2016tensorflow} natively support GPUs, but are tailored to
neural network training and inference and do not implement most of the conventional
image processing routines.
Cytokit~\cite{czech2019cytokit} implements TensorFlow-based GPU acceleration for image registration, deconvolution and quality scoring, but still relies on CPU-bound scikit-image, CellProfiler and OpenCV routines for preprocessing, segmentation and feature extraction.
PyTorch-based libraries such as Kornia~\cite{riba2020kornia} and torchvision v2~\cite{torchvision2016} offer GPU-accelerated augmentations, resampling and basic preprocessing, but are primarily designed for deep-learning workflows and lack the full spectrum of segmentation and morphometric operations.

CuPy~\cite{cupy} and RAPIDS cuCIM~\cite{cucim} represent a different approach, providing GPU-accelerated
implementations that closely mirror the APIs of NumPy, SciPy, and scikit-image.
CuPy offers a drop-in replacement for NumPy arrays and lower-level signal and image processing operations,
while cuCIM extends this paradigm to image processing routines from scikit-image.
Unlike device-agnostic solutions, these libraries require explicit data management—users
must be aware of which device their data resides on to call the appropriate CPU or GPU implementation accordingly.

Finally, several napari plugins provide GPU-accelerated routines for specific tasks. For example, the napari-accelerated pixel-and-object-classification plugin~\cite{haase2021apoc} implements OpenCL-based Random Forest pixel and object classifiers. The pycudadecon plugin~\cite{lambert2021pycudadecon} implements a CUDA-based accelerated Richardson–Lucy deconvolution~\cite{biggs1997acceleration}. However, these remain narrowly focused on their particular algorithmic domains.
The napari-pyclesperanto-assistant~\cite{clesperanto2020napari} plugin integrates clEsperanto kernels, while inheriting pyclesperanto’s API and the operation set.
The napari-cupy-image-processing plugin~\cite{haase2021napari} exposes GPU-accelerated signal and image processing rountines from CuPy and a handful of skimage-like operations within the napari ecosystem, but its coverage
remains far narrower than cuCIM’s comprehensive mapping.

%% file: sec/2_methods.tex
\section{Methods}
\label{sec:methods}

\subsection{Design principles}

\emph{cubic} implements a balanced design that seeks to address drawbacks of existing
bioimage computing tools.
We formulated the following design principles to guide the development of \emph{cubic}:

\begin{enumerate}[label=\textbf{P\arabic*}]
    \item \textbf{Comprehensive image processing operations.} Bioimage analysis workflows require diverse 
          processing steps from preprocessing to quantitative measurement, but combining multiple specialized 
          tool can be challenging. \emph{cubic} aims to provide a wide range of operations including deconvolution, filtering,
          segmentation, feature extraction, morphological operations, and image metrics commonly used in microscopy workflows. 
          This unified approach reduces dependency management and integration complexity.
    
    \item \textbf{API decoupled from GUI.} Tight coupling of the user        interface and the logic is an anti-pattern.
    \emph{cubic} exposes all core functionality through 
          programmatic interfaces, enabling seamless integration into custom pipelines and automated 
          workflows without depending on graphical interfaces. This design maximizes flexibility and 
          scriptability while requiring users to have programming expertise rather than relying on 
          point-and-click interfaces.
    
    \item \textbf{Multidimensional image support.} Modern microscopy generates complex multi-dimensional datasets 
          including 3D image volumes that many tools handle inconsistently. 
          \emph{cubic} provides native support for $2$D, $3$D, and higher-dimensional image stacks with 
          consistent APIs across all dimensionalities. This unified approach simplifies analysis of complex 
          datasets but increases implementation complexity and memory requirements for higher-dimensional 
          operations.
    
    \item \textbf{GPU acceleration.} Computationally intensive bioimage processing can be prohibitively slow 
          on CPU-only systems, especially for large 3D images. \emph{cubic} leverages GPU hardware through 
          CUDA libraries while maintaining automatic fallback to CPU implementations when GPU resources are 
          unavailable. This approach delivers significant performance improvements for supported operations 
          while maintaining broad compatibility.
    
    \item \textbf{Scientific Python integration.} Python has emerged as the dominant language for data 
          and image analysis due to its extensive ecosystem of scientific libraries, clear syntax, and 
          strong community support. \emph{cubic} ensures seamless 
          interoperability with established libraries like NumPy, SciPy, and 
          scikit-image through consistent array interfaces and data types.
    
    \item \textbf{Device agnosticism.} Hardware-specific code creates maintenance burden and
    substantial refactoring of existing codebases. \emph{cubic} enables identical code to execute 
          on both CPU and GPU hardware through unified array interfaces, allowing transparent acceleration 
          without code modification. This approach maximizes code reusability and simplifies deployment.
\end{enumerate}

\subsection{Key features}

Based on the survey of existing bioimage analysis tools in Section~\ref{sec:related_work} and our design principles, we chose to implement \emph{cubic} using SciPy~\cite{virtanen2020scipy} and scikit-image~\cite{vanderwalt2014scikit},
combined with CuPy~\cite{cupy} and RAPIDS cuCIM~\cite{cucim}.
Together, SciPy's lower-level signal and n-dimensional image processing capabilities (deconvolution, filtering)
and scikit-image's extensive higher-level image analysis operations (segmentation, morphological operations,
feature extraction) provide one of the most comprehensive collections of bioimage processing and analysis tools under a unified programmatic interface (\textbf{P1}, \textbf{P2}).
Both libraries also natively support multidimensional arrays with consistent operations across 2D and 3D data (\textbf{P3}).
Using CuPy and cuCIM as drop-in replacements for NumPy/SciPy and scikit-image enables GPU-accelerated array operations and image processing functions with mostly identical function signatures (\textbf{P4}).
Supporting highly popular SciPy/scikit-image APIs enables seamless integration with the broader NumPy ecosystem and existing scientific Python workflows through consistent array interfaces and data types (\textbf{P5}).

Although CuPy and cuCIM mostly mirror SciPy/scikit-image's API, they still require placing the input array to the correct device and importing matching device-specific functions.
We simplify this requirement by providing a device‐agnostic interface that automatically dispatches calls to the array's current device, such that unmodified code executes on both CPU and GPU (\textbf{P6}).
If a CUDA implementation is unavailable or no GPU is detected, \emph{cubic} transparently falls back to the CPU version of each routine, preserving compatibility while delivering hardware‐accelerated performance where possible.
Through this wrapping approach, \emph{cubic} provides access to the near-complete functionality of scikit-image's
comprehensive API, encompassing the broad spectrum of image processing operations including color space 
conversions, feature detection, filtering, morphological operations, segmentation algorithms, geometric 
transformations, image registration, deconvolution, and feature extraction.
By using the device-agnostic API and simply updating import statements, \emph{cubic} easily allows adding GPU acceleration to existing scikit-image code.

\emph{cubic} also implements custom algorithms not covered by existing libraries, including 
advanced image quality metrics such as Fourier Ring Correlation (FRC) and Fourier Shell Correlation 
(FSC)~\cite{nieuwenhuizen2013measuring,banterle2013fourier} for resolution assessment,
scale-invariant Peak Signal-to-Noise Ratio (PSNR)~\cite{weigert2018content} and Structural Similarity
Index Measure (SSIM)~\cite{wang2004image}, segmentation quality metrics such as Average Precision
(AP)~\cite{everingham2010pascal}, alternative implementations~\cite{tnia} of Lucy-Richardson
deconvolution~\cite{richardson1972bayesian,lucy1974iterative}, and various specialized image utility
operations for bioimage analysis, all available in both 2D and 2D. This comprehensive coverage ensures that researchers have access to
an even wider array of image processing operations needed for bioimage analysis workflows while
maintaining the device-agnostic execution model.

\subsection{Device-agnostic processing}
To illustrate the disadvantages of device-specific APIs, the snippet below shows attempts to apply a Gaussian filter to a 3D image of cell nuclei using both CPU-based scikit-image and GPU-based cuCIM implementations:

\begin{lstlisting}[style=custom]
import cupy as cp
import numpy as np
from skimage import data
import skimage.filters as filters_cpu
import cucim.skimage.filters as filters_gpu

# load example 3D image of cells
img = data.cells3d()
# select nuclei channel
img = img[:, 1] # ZYX (60, 256, 256)

# try running Gaussian filter on GPU
smooth = filters_gpu.gaussian(img)
> TypeError:
> Unsupported type <class 'numpy.ndarray'>

# move img to GPU
img = cp.asarray(img)
# now this works
smooth = filters_gpu.gaussian(img)

# try running Gaussian filter on CPU
smooth = filters_cpu.gaussian(img)
> TypeError: Implicit conversion
> to a NumPy array is not allowed.

# move img back to CPU
img = img.asnumpy()
# only now this works
smooth = filters_cpu.gaussian(img)
\end{lstlisting}

In this pattern, importing device-specific implementations of each module forces the developer to not only keep track of the current data location, but also to choose the correct device-specific implementation to match it, leading to redundant namespace clutter and cognitive overhead.  Otherwise, mismatch between the data location and function implementation triggers unexpected runtime errors. As a result, extending existing scikit-image-based bioimage analysis pipelines to run on GPU demands extensive refactoring to alternate between backends, undermining readability and maintainability.

By contrast, \emph{cubic} consolidates CPU and GPU functionality behind a single API, requiring only one import for each submodule and localizing device transfers to explicit boundary calls.  For example:

\begin{lstlisting}[style=custom]
from cubic.cuda import ascupy
from cubic.skimage import filters

# load example 3D image of cells
img = data.cells3d()
# select nuclei channel
img = img[:, 1] # ZYX (60, 256, 256)

# run Gaussian filter on CPU using skimage
smooth = filters.gaussian(img)

# transfer image to GPU
img = ascupy(img)
# run Gaussian filter on GPU using cuCIM
smooth = filters.gaussian(img) 
\end{lstlisting}

In this example, the GPU code matches exactly the widely-used scikit-image API, except for calling \verb|cubic.skimage| instead of \verb|skimage|. 
The function calls are identical, however, the operation runs on either
CPU and GPU hardware depending on the input array location.
The \verb|ascupy| function transfers the input array to the GPU, allowing
the use of GPU-accelerated implementations of the same algorithm.
The output array is also a GPU array, which can be transferred back to the CPU using \verb|asnumpy| if needed.
Persistent output location allows running existing scikit-image code without further modifications.
This design allows developers to write device-agnostic code that runs seamlessly on both CPU and GPU hardware, maximizing code reusability and simplifying deployment across different computing environments.

%% file: sec/3_results.tex
\section{Benchmarks and examples}
\label{sec:benchmarks}

We demonstrate the capabilities of \emph{cubic} using following examples:
\begin{itemize}
  \item GPU-accelerated image rescaling benchmark.
  \item Re-implementation of a CellProfiler pipeline for 3D cell monolayer segmentation.
  \item Richardson-Lucy deconvolution with the optimal iteration selection procedure guided by image quality metrics.
\end{itemize}

\subsection{3D image rescaling benchmark}

We demonstrate \emph{cubic}'s performance advantages through benchmarking image rescaling operations, comparing CPU and GPU execution across different interpolation orders and input
image sizes.

\textbf{Dataset and experimental setup.} The benchmark uses the \texttt{cells3d} dataset from scikit-image~\cite{vanderwalt2014scikit},
a 3D fluorescence microscopy image stack with dimensions 60×256×256 pixels representing a typical
confocal acquisition downscaled in XY to almost isotropic voxel size. To evaluate performance scaling with image size, we additionally test on a
2× XY-upsampled version (60×512×512 pixels), created using bilinear interpolation.

We benchmark upscaling the input image by 2× and then downscaling
the result back by 0.5× using \texttt{cubic.skimage.trasnform.rescale}, which internally uses \texttt{scipy.ndimage.zoom} with optional Gaussian filtering for anti-aliasing.
We test all interpolation orders (0-5) supported by SciPy: nearest-neighbor (order 0), linear (order 1), 
quadratic (order 2), cubic (order 3), quartic (order 4), and quintic (order 5). Anti-aliasing was enabled for downscaling operations 
to prevent aliasing artifacts.

\begin{figure}[t]
  \centering
  \includegraphics[width=\linewidth]{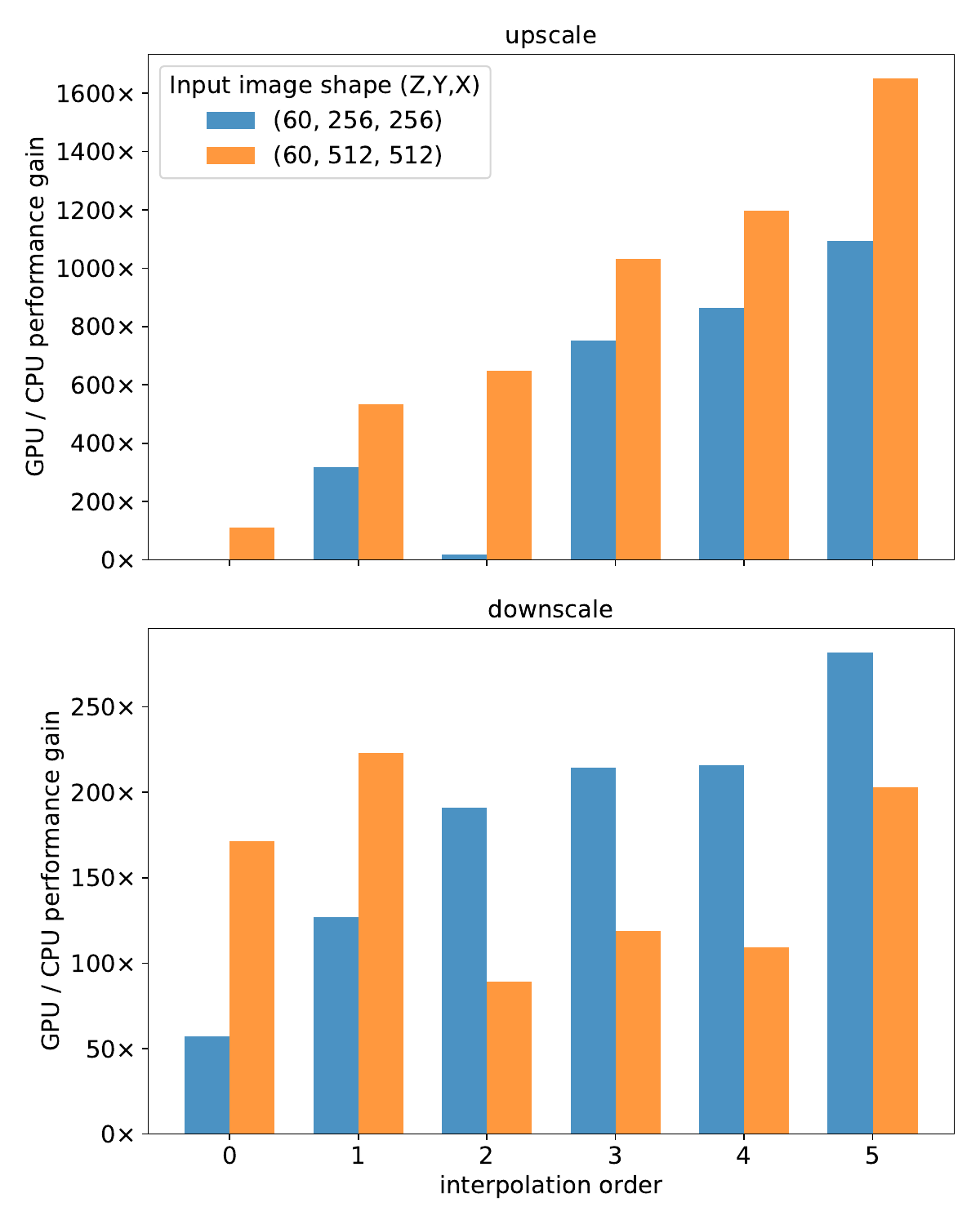}
  \caption{GPU performance gains for 3D image rescaling operations. Results show speedup factors comparing
  GPU versus CPU execution across interpolation orders (0-5) for two sizes of 3D input images.}
  \label{fig:rescale_benchmark}
\end{figure}

\begin{figure*}[t]
  \centering
  \begin{subfigure}{0.39\linewidth}
    \includegraphics[width=\linewidth]{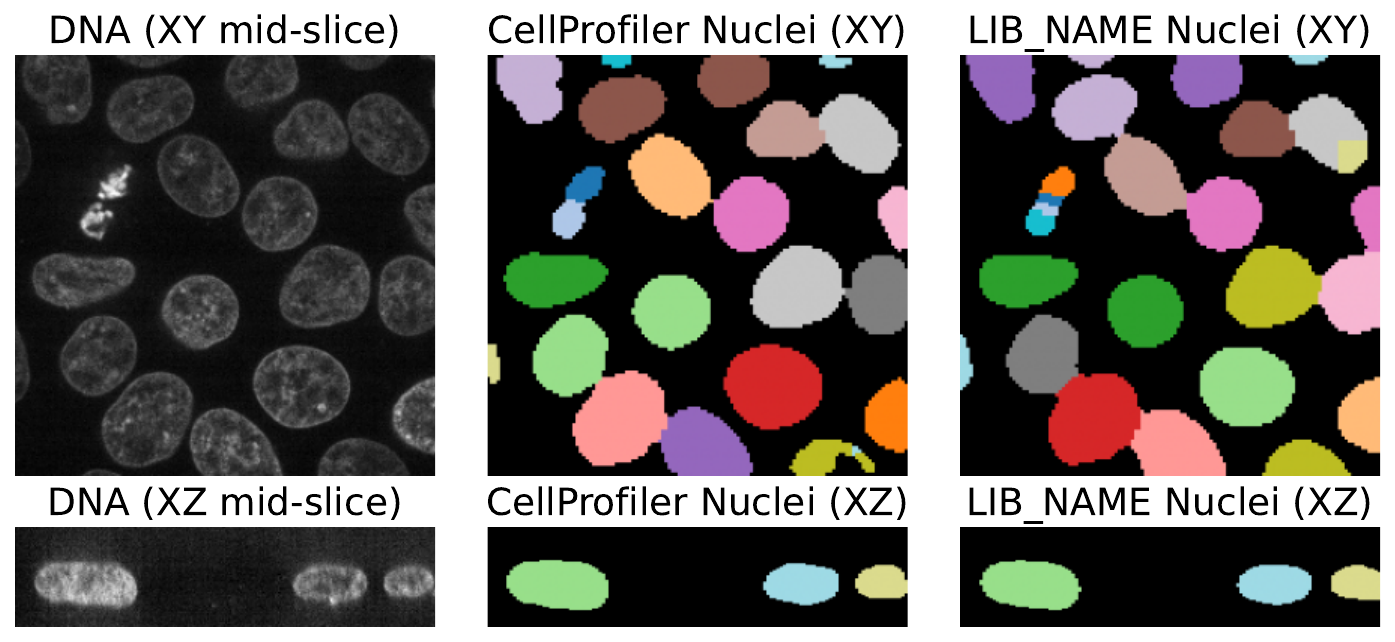}
    \caption{Visual comparison of DNA channel and nuclei segmentation results. Top row shows XY mid-slice views, 
    bottom row shows XZ mid-slice views. From left to right: original DNA channel, CellProfiler nucleus 
    segmentation, and \texttt{\emph{cubic}} nucleus segmentation.
    \emph{cubic} implementation produces segmentation results similar to those by the original CellProfiler pipeline (differences are due to CellProfiler internally setting some default parameter values differently from those in scikit-image without exposing them in the GUI).}
    \label{fig:3d_monolayer_visual}
  \end{subfigure}
  \hfill
  \begin{subfigure}{0.59\linewidth}
    \includegraphics[width=\linewidth]{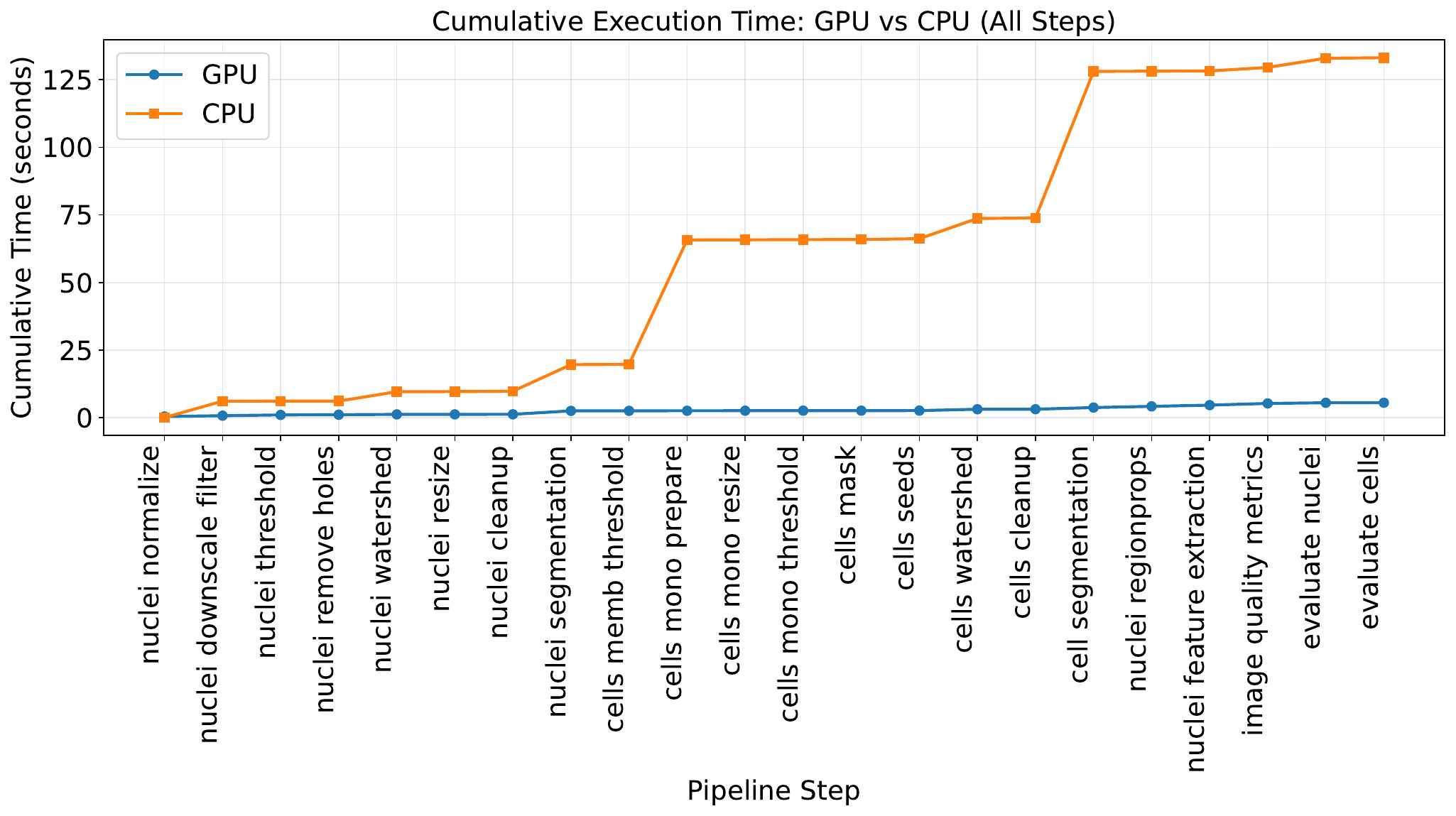}
    \caption{Cumulative execution time comparison between GPU and CPU implementations.}
    \label{fig:3d_monolayer_timing}
  \end{subfigure}
  \caption{CellProfiler 3D monolayer segmentation benchmark results. (a) Visual validation showing similar 
  segmentation quality between CellProfiler and \emph{cubic} implementations. (b) Performance comparison 
  demonstrating significant computational speedup using GPU.}
  \label{fig:3d_monolayer_results}
\end{figure*}

\textbf{Results.} We found substantial GPU acceleration across
all tested configurations (Figure~\ref{fig:rescale_benchmark}). For upscaling operations, GPU speedup ranges from approximately 10× (order 0)
for the original image size to over 1600× (order 5) for the larger upsampled input, with higher-order
interpolation methods showing dramatically greater acceleration due to their increased computational
complexity.
Downscaling operations show more variable but still substantial performance improvements,
ranging from 50-300× speedup across different interpolation orders.
The performance scaling with input image size confirmes the known advantage of GPU acceleration for
larger images, which consistently achieves a speedup of a least 100× across both operations and both input sizes.
Even the simplest nearest-neighbor interpolation (order 0) shows significant speedups of 10-170×,
demonstrating that using \emph{cubic} can provides substantial computational benefits across the
entire spectrum of image rescaling operations.

\subsection{CellProfiler 3D monolayer segmentation}

We demonstrate \emph{cubic}'s capabilities by reproducing a CellProfiler pipeline for 3D monolayer 
segmentation, while also comparing performance between CPU and GPU execution.
Notably, CellProfiler v4~\cite{stirling2021cellprofiler4} relies on scikit-image for most of its core image processing operations, making our GPU-accelerated \emph{cubic} wrapper a natural performance enhancement for such workflows.
Note, that the goal of this example was not to perfectly reproduce CellProfiler pipeline, but to use similar operations to evaluate acceleration allowed by the use of GPU computation.

\textbf{Dataset description.} The experiment uses a 3D fluorescence microscopy image of a cell monolayer from the
image set BBBC034v1 Thirstrup et al. 2018, available from the Broad Bioimage Benchmark Collection~\cite{ljosa2012annotated}.
The 3D image stack has three channels: membrane (channel 0), mitochondria (channel 1), and DNA (channel 2).
The images have pixel dimensions of 0.065 $\mu$m in X and Y, with 0.29 $\mu$m Z-spacing, representing typical
confocal microscopy acquisition parameters for monolayer studies.

\textbf{CellProfiler pipeline.} The pipeline performs hierarchical segmentation to identify: (1) individual cell 
nuclei from the DNA channel, and (2) whole cells using membrane signal constrained by nuclear seeds. This 
two-step approach mirrors common workflows in cell biology where nuclear segmentation provides reliable seeds 
for subsequent cell boundary detection.

The reference CellProfiler implementation follows a multi-stage approach:
\begin{itemize}
    \item \textit{Nuclei segmentation}: DNA channel normalization, downscaling (0.5$\times$), median filtering 
    (ball radius 5), Otsu thresholding, hole filling (area threshold 20), watershed segmentation (ball radius 10), 
    upscaling, and size filtering (minimum 50 pixels).
    \item \textit{Cell segmentation}: Multi-Otsu thresholding of membrane channel (3 classes), hole removal, 
    monolayer mask creation from combined channels with morphological closing (disk radius 17) at 0.25$\times$ 
    resolution, seed generation from eroded nuclei (ball radius 5), and watershed segmentation (ball radius 8) 
    with size filtering (minimum 100 pixels).
\end{itemize}

\textbf{\emph{cubic} implementation.} Our implementation reproduces this pipeline using GPU-accelerated \emph{cubic} functions, maintaining identical parameter values and processing steps where possible.
Additionally, we evaluated segmentation quality against CellProfiler results by calculating average precision (AP) on labeled masks.
To both demonstrate performance on standard image-quality measures and tie them to segmentation agreement, we additionally computed peak signal-to-noise ratio (PSNR) and structural similarity index (SSIM) on intensity images masked with predicted segmentation labels. When evaluated within the masks, higher PSNR/SSIM values imply closer spatial agreement of segmentation results.
The modular design allows easy switching between CPU and GPU execution while preserving numerical accuracy.

\begin{figure*}[t]
  \centering
  \includegraphics[width=\linewidth]{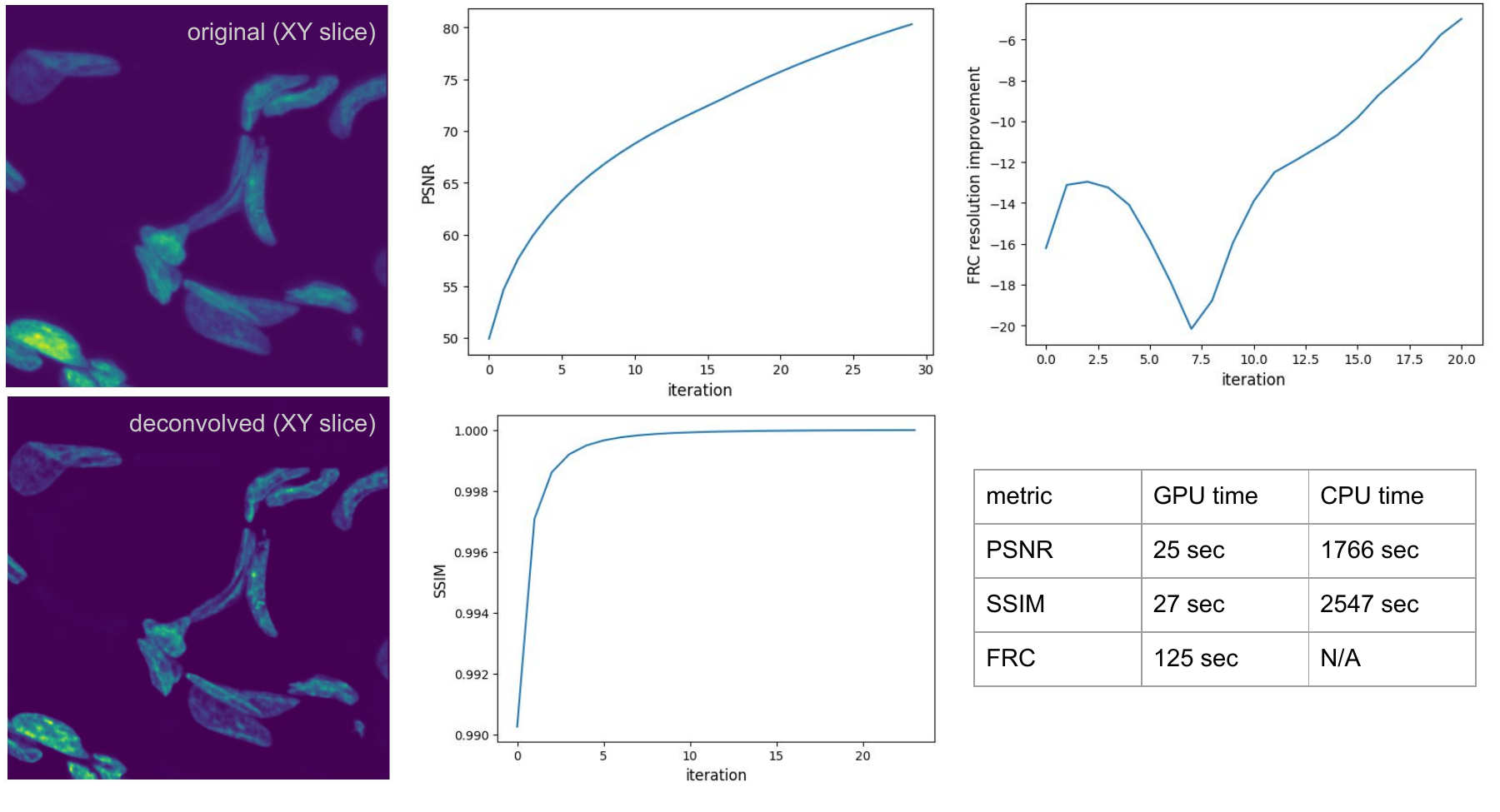}
  \caption{Richardson-Lucy deconvolution of a large 3D image volume ($30\times2160\times2560$) with per-iteration image quality metrics tracking and CPU vs GPU performance comparison.}
  \label{fig:deconv_visual}
\end{figure*}

\textbf{Results.} Figure~\ref{fig:3d_monolayer_visual} shows visual comparison between CellProfiler and 
\emph{cubic} segmentations in both XY and XZ views, demonstrating high fidelity reproduction of the original pipeline.
Performance analysis (Figure~\ref{fig:3d_monolayer_timing}) reveals substantial speedup on GPU, with total 
pipeline acceleration of 25$\times$ compared to CPU execution (5.62 vs 133.43 sec).
Individual steps show varying degrees of 
improvement, with morphological operations and image filtering achieving the largest gains due to their parallel 
nature.

This example demonstrates the potential ability to replace existing CellProfiler workflows that internally use scikit-image operations to provide significant computational advantages for high-throughput applications.

\subsection{Richardson-Lucy deconvolution}

We demonstrate \emph{cubic}'s advanced deconvolution capabilities through Richardson-Lucy deconvolution 
with optimal iteration selection guided by image quality metrics, where both deconvolution and metric calculation happen on the same device.

\textbf{Dataset description.} The experiment uses a single 3D image stack of Hoechst-stained
astrocyte nuclei acquired with a Yokogawa CQ1 confocal microscope~\cite{kalinin2025foreground}.
Theoretical 3D point spread functions for each individual image volume were modeled using the Richards and Wolf
algorithm~\cite{richards1959electromagnetic} from the PSFGenerator plugin~\cite{kirshner20133} for Fiji.
Both the image and the PSF have the same ZYX size of $30\times2160\times2560$.

\textbf{Richardson-Lucy algorithm.} The Richardson-Lucy algorithm is an iterative deconvolution method that 
maximizes likelihood under Poisson noise assumptions. Each iteration involves forward convolution with the 
PSF, ratio computation with the observed image, back-convolution with the flipped PSF, and multiplicative 
update of the estimate. The algorithm's effectiveness depends critically on selecting the optimal number 
of iterations to balance noise suppression and detail preservation.

Traditional Richardson-Lucy implementations require manual iteration 
count selection, often leading to under- or over-deconvolution. We implemented automatic stopping 
criteria that monitor image quality improvement between consecutive iterations. The threshold indicates when further iterations provide diminishing returns, enabling automated optimal 
stopping without user intervention.

\textbf{\emph{cubic} implementation.} Our GPU-accelerated implementation leverages \emph{cubic}'s FFT 
operations and element-wise arithmetic to achieve substantial speedup over CPU-based approaches. Key 
optimizations include:
\begin{itemize}
    \item GPU-resident FFT operations for convolution steps
    \item Memory-efficient in-place operations to minimize data transfer
    \item Vectorized FRC computation using GPU-accelerated cross-correlation
    \item Automated convergence monitoring with configurable thresholds
\end{itemize}

\textbf{Evaluation metrics.} We assess deconvolution quality using multiple metrics: (1) peak signal-to-noise 
ratio (PSNR) and structural similarity index (SSIM) against ground truth, (2) single-image
Fourier Ring Correlation-derived resolution~\cite{koho2019fourier}.

\textbf{Results.} Figure~\ref{fig:deconv_visual} demonstrates results of metric-guided iteration selection, with different metrics capturing various image quality changes during the deconvolution process.

Performance benchmarks (Figure~\ref{fig:deconv_visual}) reveal GPU acceleration of ~50$\times$ 
depending on the metric, with the FRC taking too long to measure on CPU.

This example showcases \emph{cubic}'s capability to implement sophisticated, research-grade algorithms 
with both computational efficiency and methodological rigor, making advanced deconvolution accessible for 
high-throughput microscopy applications.

\section{Discussion}
\label{sec:conclusion}
In this study we introduced \emph{cubic}, a lightweight Python library that unites standard SciPy and scikit-image routines and their GPU-accelerated alternatives in CuPy and cuCIM. Specficially, \emph{cubic} retains the exact function signatures and patterns familiar to users of scikit-image and scipy.ndimage; switching a pipeline to GPU execution simply requires replacing imports and a single array transfer call. This lowers the barrier to entry for high-throughput, reproducible bioimage analysis and ensures that researchers can scale their 2D and 3D workflows across heterogeneous computing environments without sacrificing readability or maintainability.
Through example benchmarks on representative 3D microscopy pipelines—covering 3D smoothing, multi-step segmentation and feature extraction, and deconvolution—we demonstrated speed-ups of 10–1500× over CPU-only workflows in each major processing stage.

\section*{Acknowledgements}
This work was partially supported by Chinese Key-Area Research and Development Program of Guangdong Province (2020B0101350001).
This work was partially supported by the Human Frontier Science Program (RGY0081/2019 to S.S.) and a grant from the National Institutes of Health NIGMS (R35 GM122547 to A.E.C.).
Xin Rong of the University of Michigan donated NVIDIA TITAN X GPU used for this research, and the NVIDIA Corporation donated the TITAN Xp GPU used for this research.